\documentclass[11pt,a4paper]{article}
\usepackage[hyperref]{emnlp2018}
\usepackage{times}
\usepackage{latexsym}
\usepackage{url}

\usepackage[normalem]{ulem}

\newcommand{\anonref}[1]{#1}
\usepackage[utf8]{inputenc}
\usepackage[color=yellow]{todonotes}
\usepackage[small,bf]{caption}

\usepackage{multirow}

\aclfinalcopy 

\title{Exploring Gap Filling as a Cheaper Alternative to Reading
  Comprehension Questionnaires when Evaluating Machine Translation for
  Gisting}

 \author{Mikel L.\ Forcada \\ \\
   Dept.\ Lleng.\ i Sist.\ Inf., \\
   Universitat d'Alacant \\
   E-03690 St.\ Vicent del Raspeig, \\
   Spain \\
   {\tt mlf@ua.es} \\\And
   Carolina Scarton, \\ \textbf{Lucia Specia} \\
   Dept.\ of Comput.\ Sci., \\
   Univ.\ of Sheffield,  \\
   Sheffield S1 4DP, U.K \\
   {\tt \{l.specia,c.scarton\}}\\{\tt @shef.ac.uk} \\\And
   Barry Haddow, \\ \textbf{Alexandra Birch} \\
   School of Computing, \\
   Univ.\ Edinburgh, \\
   Edinburgh EH8 4AB, U.K \\
   {\tt \{bhaddow,a.birch\}}\\ {\tt@ed.ac.uk}\\
 }

\date{}

\begin{document}
\maketitle
\begin{abstract}
A popular application of machine translation (MT)
  is 
  \emph{gisting}: MT is consumed \emph{as is} to make sense of text in
  a foreign language. Evaluation
  of the usefulness of MT for gisting is surprisingly
  uncommon. The classical method uses \emph{reading comprehension
    questionnaires} (RCQ), in which informants are asked to answer
  professionally-written 
  questions 
  in their language 
  about a foreign
  text that has been machine-translated into their language. 
  Recently,
  \emph{gap-filling} (GF), a form of \emph{cloze} testing, has been
  proposed as a cheaper alternative to RCQ. In
  GF, 
  certain words are removed from reference 
  translations and readers are asked to fill the gaps left using the
  machine-translated text as a hint.  This paper reports, for the
  first time, a comparative evaluation, using both RCQ and GF, of 
  translations from multiple MT systems for the same foreign texts, and a systematic study
  on the effect of variables such as gap density,
  gap-selection strategies, and document context in GF.
  %
  The main findings of the study are: (a) both RCQ and GF clearly
  identify MT to be useful; (b) global RCQ and GF rankings for the
  MT systems are mostly in agreement; (c) GF scores vary very widely across informants, making comparisons among MT systems hard,
and (d) unlike RCQ, which is
  framed around documents, GF evaluation can be
  framed at the sentence level.
  These findings support the use of GF as a cheaper alternative to RCQ.
\end{abstract}

\section{Introduction}

\subsection{Machine translation for gisting}

Machine translation (MT) applications fall in two main groups:
\emph{assimilation} or \emph{gisting}, and
\emph{dissemination}. Assimilation 
refers to the use of the raw MT output
to make sense of 
foreign texts.
Dissemination refers to the use of the MT output as a draft
translation that can be 
\emph{post-edited} 
into a publishable translation.  The needs of both groups of
applications are quite different; for instance, an otherwise
\emph{perfect} Russian to English translation but with no articles
(\emph{some}, \emph{a}, \emph{the}), is likely to be fine for
assimilation, but would need substantial post-editing for
dissemination.
State-of-the-art MT systems are however usually evaluated ---even if
manually--- (and optimized) with respect to their ability to produce
translations that resemble references,
regardless of the intended
application for the system.

Assimilation is by far the main use of MT in number of words
translated.
It is either explicitly invoked, for instance, by visiting webpages
such as Google Translate, 
or integrated into browsers and social networks.  
Raw MT may sometimes be
the only feasible 
option,\footnote{Twenty-five years ago,
  \cite[p.~261]{sager93b} already hinted at MT-only scenarios:
  ``there may, indeed, be no single situation in which either human or
  machine would be equally suitable.''} 
for instance when dealing with
user-generated content or ephemeral
material (such as product descriptions in e-commerce).

\subsection{Evaluation of MT for gisting} 
A straightforward
(but costly) way to evaluate
MT for gisting
\emph{measures} the performance of target-language readers in a
text-mediated task ---for instance, a software installation task
\cite{castilho2014does}--- by using raw MT and compares it with the
performance reached using a professional translation of the text.

However, there may be 
scenarios 
without an obvious associated task: news, product and service reviews,
or literature. On the other hand, even with a clear associated task,
task completion evaluation is also quite expensive.  It is therefore
desirable to have alternative objective indicators which work as good surrogates
for actual task-oriented success.

Some authors have proposed eye-tracking
\cite{Doherty2009,Doherty2010,Stymne2012,Doherty2014,castilho2014does,Klerke2015,Castilho2016,Sajjad2016}
as a measure of machine translation usefulness, but the technique is
expensive and the evidence gathered is rather indirect and does not
have a straightforward interpretation in terms of usefulness.

There are many methods in which informants are asked to \emph{judge}
the \emph{quality} of machine-translated sentences, usually as regards
their monolingual \emph{fluency} (nativeness, grammaticality), 
their bilingual \emph{adequacy} (how much of the information in the
source sentence is present in the machine-translated sentence), or even monolingual adequacy (how much of the information in the
reference sentence is present in the machine-translated sentence);
informants may be asked either to \emph{directly assess} MT outputs by giving
values to these indicators in a predetermined scale or to \emph{rank}
a number of MT outputs for the same source sentence (sometimes being
asked to consider aspects such as adequacy, fluency, or both).
Direct assessments of adequacy and MT ranking are the official evaluation procedure for the most recent WMT translation shared task campaigns \citep{bojar2016findings,bojar2017findings}.
Other researchers use post-task questionnaires \cite{Stymne2012,Doherty2014,Klerke2015,Castilho2016} to assess the perceived usefulness of MT output.


Direct assessment, ranking or
  post-task questionnaire evaluation methods are clearly subjective
and require informants to make ``in vitro'' \emph{judgements} about
the \emph{quality} of MT outputs, without considering their
\emph{usefulness} for a specific ``in vivo'', real-world application.




\subsection{Reading comprehension questionnaires}
Reading comprehension questionnaires (RCQ), as used in the assessment
of foreign-language learning, are the standard approach
to evaluate MT for gisting that measures reader performance in
response to MT. 
Readers
answer questions using either a machine-translated or a
professionally-translated version of the source text and their
performance on the tests (i.e.\ to what extent they answer questions
correctly) using the two sets of texts is then compared.  RCQ are
however quite costly: a human translation is needed for a control
group and questions
need to be 
professionally written and
often 
manually marked. 

%

RCQ has a long history as an MT evaluation method. \citet{Tomita1993},
\citet{Fuji1999}, and \citet{Fuji2001} evaluate the
\emph{informativeness} or \emph{usefulness} of English--Japanese MT by
using standardized English-as-a-foreign-language RCQs (TOEFL, TOEIC)
which have been machine translated into Japanese and they are
sometimes capable of distinguishing MT systems. \citet{Jones2005a},
\citet{jones2005measuring}, \citet{jones2007ilr}, and
\citet{jones2009machine} use the structure of standardized language
proficiency tests (Defence Language Proficiency Test, Interagency
Language Roundtable) to evaluate the readability of Arabic--English MT
texts. MT'ed documents are found to be harder to understand than
professional translations, and that they may be assigned an
intermediate level of English proficiency. \citet{berka2011quiz}
collected a set of English short paragraphs in various domains,
created yes/no questions in Czech about them, and machine translated
the English paragraphs into Czech with different MT systems. They
found that outputs produced by different MT systems lead to different
accuracy in the annotators' answers.
\citet{weiss2012error} evaluate comprehension of Polish--English translations using RCQ tests and found that a text with more MT errors have less correct answers than a text with fewer MT errors.
Finally, \citet{Stymne2012} use RCQ to validate eye-tracking as a tool for MT error analysis for English--Swedish.
Interestingly, for one of their systems, the number of correct answers in the RCQ tests were higher than for the human translation.
However, test takers were more confident in answering questions about the human translations than about the MT outputs.

In this paper we explore RCQ as a measure of MT quality by using the
CREG-mt-eval corpus \cite{scarton16p}. In contrast to previous work,
this paper presents an evaluation of MT quality based on open
questions that have different levels of difficulty (as presented in
Section \ref{se:method}) for a considerable amount of documents ($36$ in contrast to only $2$ analysed by \citet{weiss2012error}).

\subsection{An alternative: evaluation via gap-filling}
An alternative approach to RCQs, 
\emph{gap filling} (GF), 
has been recently proposed
\cite{trosterud2012evaluating,oregan2013p,ageeva2015evaluating,jordan17p} 
based on another typical way of measuring reading comprehension:
\emph{cloze} (or \emph{closure}) testing \cite{taylor1953cloze}.
Instead of a question, readers get an incomplete sentence with one or
more words 
replaced by gaps, and are asked to fill the gaps.
Indeed, GF may be seen as equivalent
to
the answering of simple reading comprehension questions: for instance, a
question like \emph{Who was the president of the Green Party in 2011?}
would be equivalent to the sentence with one gap \emph{In 2011,
  \_\_\_\_\_\_\_\_\_ was the president of the Green Party.}

GF tasks are prepared by automatically punching gaps in
\emph{reference} sentences 
taken from a
professional translation of the source text.
Informants are given the machine-translated sentence as a ``hint'' for the gap-filling task; therefore,
we may view GF as a way of automatically generating questions to evaluate the MT
output. The evaluation measure is the proportion of gaps that can be successfully
filled using MT as a hint. This can be compared with the success rate in 
the case where no hint (MT) is provided, to give an estimate of the usefulness of MT output.

Note that \emph{cloze} testing evaluation of machine translation was
attempted decades ago in a completely different \emph{readability}
setting: gaps were then punched \emph{in machine-translated output}
and informants tried to complete them without any further hint
\cite{crook65tr,sinaiko72j}. This work was reviewed and extended later
by Somers and Wild \shortcite{somers00p}. But filling gaps in
machine-translated output may be unnecessarily challenging and
therefore make evaluation less adequate: for instance, informants
would sometimes have to fill gaps in disfluent or ungrammatical text,
which is much harder than filling them in a fluent, professionally
translated reference, or, even in fluent output, a crucial content
word that has been removed may be very hard to guess unless the
surrounding text is very redundant. Moreover, the GF method described
here has an easier interpretation in terms of its analogy to RCQ.

This paper systematically builds upon previous work on GF to obtain
experimental evidence that gap-filling is a viable, lower-cost
alternative to RCQ evaluation. Its main {\bf contributions} are:
\begin{itemize}
\item While Trosterud and Unhammer
  \shortcite{trosterud2012evaluating}, O'Regan and Forcada
  \shortcite{oregan2013p}, and Ageeva et al.\
  \shortcite{ageeva2015evaluating} used GF just to demonstrate the
  usefulness of a single rule-based MT system for each language pair
  studied, this paper, like Jordan et al.'s \shortcite{jordan17p},
  performs a comparison of several MT systems for the same language
  pair.
\item Previous work
  \cite{trosterud2012evaluating,oregan2013p,ageeva2015evaluating,jordan17p}
  simply assumes the validity of GF as an evaluation method for MT
  gisting, in some cases arguing about its equivalence to RCQ. Ours is the first work to actually compare GF and RCQ evaluation
  of the same MT systems.
\item Previous work used sentences
  \cite{trosterud2012evaluating,oregan2013p,ageeva2015evaluating} or
  short excerpts of text \cite{jordan17p}, but did not study the
  influence of a larger, document-level machine-translated context around the
  target sentence, as it is done here.
\item This paper explores for the first time a
  gap-positioning strategy based on an approximate computation of gap
  entropy, and compares it to random placing of gaps. 
\end{itemize}


The paper is organized as follows: section~\ref{se:method} describes
the design and implementation of both evaluation methods, RCQ and GF;
then section~\ref{se:results} reports and discusses the results
obtained; and, finally, concluding remarks (section~\ref{se:concluding}) close
the paper.

\section{Methodology}
\label{se:method}

\subsection{Data and informants}

We use an extended version of CREG-mt-eval
\cite{scarton16p}, a version of the expert-built CREG reading comprehension corpus
\cite{Ott2012} for 2nd-language learners of 
German.
CREG
was
originally created 
to build and evaluate systems that
automatically correct answers to open questions.  
CREG-mt-eval contains 108 source (German) documents
with
different domains, including literature, news, job adverts, and others
(on average 372 words and 33 sentences per document). The
original documents were machine-translated in December 2015 into
English using four systems: an in-house baseline\footnote{\url{http://www.statmt.org/moses/?n=moses.baseline}} statistical
phrase-based Moses \cite{Kohen2007} system trained on WMT 2015 data \cite{bojar2015findings}, Google
Translate,\footnote{\url{http://translate.google.co.uk/}, presumably a
  statistical system at that time.}
Bing\footnote{\url{https://www.bing.com/translator/}, also presumably
  a statistical system at that time.} and
Systran.\footnote{\url{http://www.systransoft.com/}, presumably a hybrid rule-based / statistical system at that time. }
CREG-mt-eval also contains professional translations of a subset of
36 documents (90--1500 words) as a control group to check whether
the questions are adequate for the task. 
All questions from the CREG original questionnaires (in German) were
professionally translated to English. On average, there are 8.8 questions per document.

The questions in CREG-mt-eval 
are classified \cite{Meurers2011} as:
{\em literal}, when they can be answered directly from the text 
and refer to explicit knowledge, such as names, dates (79\% of the total number of questions);
{\em reorganization}, also based on literal text
understanding, but %
requiring the combination of
information from
different parts of the 
text  (12\% of the total number of questions); and {\em inference}, which 
involve combining literal information with world
knowledge  (9\% of the total number of questions). 

Following \anonref{\citet{scarton16p}}, 
test takers (informants) for both GF and RCQ were fluent
English-speaking volunteers, 
staff and students at the University of Sheffield,
who were paid (with a 10~GBP online gift certificate) to complete the task. 

\subsection{Reading comprehension questionnaire task}

For the version of CREG-mt-eval used herein, thirty informants 
were given a set of six documents each and answered three to five
questions per document, using only the English document (either
machine- or human-translated) provided.
Therefore, for each of the 36 original documents, questions were answered using each machine translation system or the human translation. 
Each document was only evaluated by one informant. 
The original German document was not given. 
%
The guidelines were similar to those used in other reading
comprehension tests: test takers were asked to answer the questions
based on the document provided. They were also advised to read the
questions first and then look for the information required on the text in
order to speed up the task. 
%
%
Questions in CREG-mt-eval were marked as
proposed by Ott et al.~\shortcite{Ott2012}: {\em correct answer} (1 mark), if the answer is  correct and complete; 
{\em extra concept} (0.75 marks), when incorrect
additional concepts are added; {\em missing concept}
(0.5 marks), when important concepts are missing;
{\em blend} (0.25 marks) when there are both extra and missing concepts;
 and {\em incorrect} (0 marks), when the answer is
incorrect or missing. 

Given the marks and the type of question, RCQ overall scores ($f$) are calculated as:   
\[
f = \alpha \cdot \frac{1}{N_l}\sum_{k=1}^{N_l}{l_k}+ \beta \cdot \frac{1}{N_r}\sum_{k=1}^{N_r}{r_k}+ \gamma \cdot \frac{1}{N_i}\sum_{k=1}^{N_i}{i_k},
\]
where $N_l$, $N_r$ and $N_i$ are the number of literal, reorganization and inference questions, respectively, $l_k$, $r_k$ and $i_k$ are real values between $0$ and $1$, 
according to the mark of question $k$, and $\alpha$, $\beta$ and $\gamma$ are weights for the different types of questions.

We experiment with three different types of scores: 
{\em simple} (same weight for all question types: $\alpha = \beta = \gamma = 1.0$), 
i.e. marks are averaged giving all questions the same importance; 
{\em weighted}, i.e. marks are
averaged using different weights for different types of question
($\alpha = 1$, $\beta = 2$ and $\gamma = 3$);\footnote{These values reflect the expected relative difficulty of questions: inference harder than reorganization, and reorganization harder than literal.} and
{\em literal}, where only marks for literal questions are used 
to compute the average quality score ($\alpha = 1$, $\beta = \gamma = 0$). The last score is interesting because literal questions are the most similar to gap-filling problems and correspond to almost 80\% of the corpus and they should be easier to answer than other types. Therefore, problems in answering a literal question may be a sign of a bad quality translation.

\begin{figure*}[ht]
  \centering
  \includegraphics[scale=0.6]{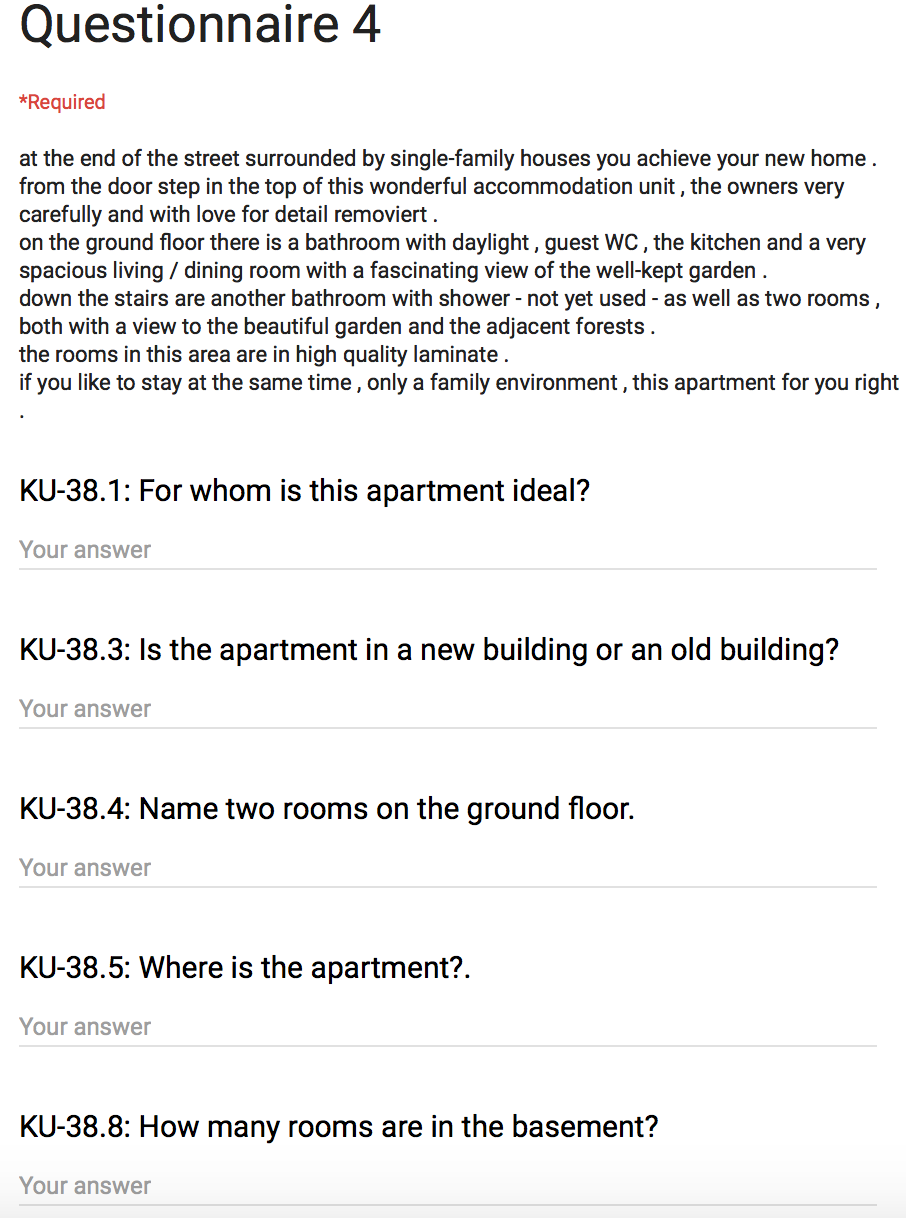}
  \caption{A screenshot of a RCQ questionnaire.}
  \label{fig:RCQ}
\end{figure*}

Figure \ref{fig:RCQ} shows an example of the questionnaires presented to the test takers. In this example, the first, second and last questions are inference questions, whilst the third and fourth questions are literal questions.

\subsection{Gap filling task}

Twenty different kinds of configurations were used in problems posed
to informants. Sixteen configurations used the four MT systems to
generate hints, in two modalities (showing the full machine-translated
document, or just the problem sentence) and with two different gap
densities (10\% or 20\%). We added 4 additional configurations with no
hint, using the same two gap densities, and with two different
gap-selection strategies (statistical language model entropy and
random).

The gap entropy at position $k$ of sentence $w_1^N$ is given by,
\[
H(k,w_1^N)= - \sum_{x\in V} p(x|w_1^{n},k) \log_2 p(x|w_1^{n},k),
\] with $V$ the target vocabulary
(including the unknown word UNK), and with
\[p(x|w_1^{n},k) =\frac{p(w_1^{k-1} x w_{k+1}^N)}{\displaystyle \sum_{x'\in V}
p(w_1^{k-1} x' w_{k+1}^N)},\]
estimated using a 3-gram language model trained trained using
KenLM
\cite{heafield2011kenlm} on the English NewsCommentary version 8
corpus.\footnote{\protect\url{http://www.statmt.org/wmt13}}
Gaps are punched in order of decreasing entropy, 
disallowing gaps at stop-words or punctuation,
 and ensuring that two gaps are never consecutive or separated only by
stop-words or punctuation. 

To select important sentences for the test, for each of the reference documents,
the best single-sentence summary was selected as the problem sentence
using
GenSim.\footnote{\protect{\url{https://rare-technologies.com/text-summarization-with-gensim/}}; the percentage of text to be kept in the summary is reduced until it contains a single sentence.}

Each of 60 informants 
was given
exactly one problem per document.
Problem configurations were assigned such that each informant tackled at
least one problem in each configuration, and each document was evaluated 3 times in each
configuration. The mean time per problem was about
1 minute. 

\begin{figure*}[ht]
  \centering
  \includegraphics[scale=0.42]{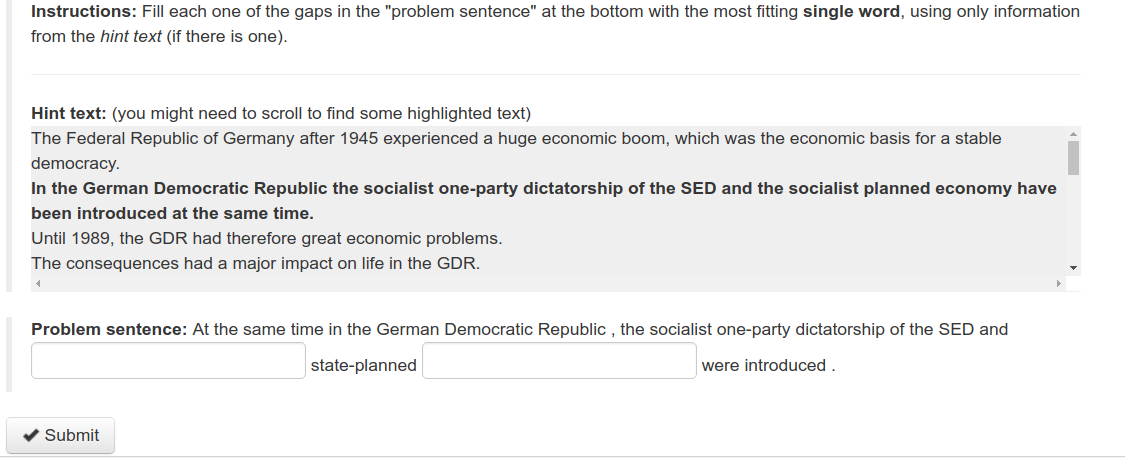}
  \caption{A screenshot of the gap-filling evaluation interface, showing a whole machine-translated document as a hint (with the key sentence highlighted).}
  \label{fig:GFSS-doc}
\end{figure*}

To create the user interface for the task we modified\footnote{\url{https://github.com/mlforcada/Appraise}} 
Ageeva et al.'s~\shortcite{ageeva2015evaluating}
version 
of an older version (2014) 
of 
\citeauthor{mtm12_appraise}'s
(\citeyear{mtm12_appraise})
Appraise.\footnote{\url{https://github.com/cfedermann/Appraise}}
Each problem was presented in Appraise in a
single screen, divided in three sections. The top of each screen
reminded informants about the objective of the task. 
Immediately below, a machine-translated \emph{Hint text} is provided for those 16 configurations that
have one. The sentence in the hint text corresponding to the problem sentence
is highlighted when a complete document is provided.  At the bottom of
the screen, the \emph{Problem sentence} 
containing the gaps to be filled 
is provided. Figure~\ref{fig:GFSS-doc} shows a screenshot of the interface, where a whole machine-translated document is shown as a hint, with the key sentence highlighted.
The score for each problem and configuration is
simply the ratio of correctly filled gaps.

\section{Results}
\label{se:results}


\begin{table*}[ht!] 
  
    \begin{center}
    \begin{tabular}{c | c | c | c | c | c | c | c | c}

      & BLEU & NIST & \multicolumn{3}{c|}{RCQ scores} &  \multicolumn{3}{c}{GF scores}\\ \hline
       &  &  & Simple & Weighted & Literal & Overall & 10\% & 20\% \\
      \hline
 	 Google & $\mathbf{0.306} $ & $\mathbf{4.66}$ & $\mathbf{0.753}$ & $\mathbf{0.748}$ & $0.776$ & $0.592$ & $0.565$ & $0.619$ \\     
      Bing &  $0.281$ & $4.40$ & $0.709$ & $0.695$ & $0.734$ & $\mathbf{0.618}$ & $\mathbf{0.595}$& $\mathbf{0.640}$ \\ 
      Homebrew & $0.241$ & $4.51$ & $0.594$ & $0.577$ & $0.608$ & $0.550$ & $0.547$ & $0.553$ \\
      Systran & $0.203$ & $3.05$ & $0.680$ & $0.670$ & $0.701$ & $0.569$ & $0.544$ & $0.595$ \\\hline
      MT Average & & & $0.684$ & $0.673$ & $0.705$  & $0.582$ & $0.563$ & $0.602$ \\\hline	
      Human & $1.000$ & $10.0$ & $0.813$ & $0.810$ & $0.872$ \\\hline
      No hint  (random) & \multicolumn{5}{c|}{}   & $0.258$   & $0.302$ & $0.213$ \\
      No hint  (entropy) &  \multicolumn{5}{c|}{} & $0.193$   & $0.191$ & $0.195$ \\
      No hint (average)  & \multicolumn{5}{c|}{}  & $0.225$   & $0.247$ & $0.204$ \\\hline
    \end{tabular}
    \end{center}
  
    \caption{A comparison of BLEU and NIST scores, RCQ marks in the three possible weightings, and GF success rates at different densities.}
    \label{tb:quickcomp}
 \end{table*}

 Table \ref{tb:quickcomp} shows, for each system, the averaged
 informant performance  (see Appendix~\ref{app:supplemental} for details) for the GF and RCQ quality scores explained
 previously; BLEU and NIST scores are also given as a
 reference. In view that score distributions are actually very far
 from normality, the usual significance tests (such as Welch's
 $t$-test) are not applicable; therefore, statistical significances of
 differences between RCQ and GF scores will be reported throughout
 using the distribution-agnostic Kolmogorov--Smirnov
 test.\footnote{\protect{\url{https://en.wikipedia.org/wiki/Kolmogorov-Smirnov_test}}}
 Note that previous work in RCQ
 did not provide statistical significance when comparing different
 hinting conditions, and that only Jordan et al.\ \shortcite{jordan17p}
 provided that information for GF.



\subsection{Reading comprehension questionnaire scores}

According to all three variations of RCQ scores, and
contrary to BLEU and NIST, Systran appears to be better than the homebrew
Moses.
The RCQ scores for the professionally translated documents ('Human' row 
on the table)
are higher than those for the best MT system,
which shows that the questions are answerable from the texts and that informants did follow the guidelines
as expected. 

We also report the statistical significance of
score differences and find (a) the only statistically significant
difference at $\alpha<0.05$ between MT systems for any score type is
between Google and the homebrew Moses; (b) all three scores of Bing,
Google and Systran are statistically indistinguishable among them; 
(c) some (but not all) scores obtained with the professional
translation are not statistically different from those obtained with
Google, Bing or Systran MT output;  and (d) all three scores obtained with
the professional translation are statistically distinguishable from
those with Moses output. 


\subsection{Gap-filling}

\paragraph{Gap placement strategy:}
Filling of gaps in the absence of a hint was done in two
configurations: one where gaps were punched at random, and one where
gaps were punched where LM entropy was maximum.   Entropy appears to
make gap filling more difficult in the absence of hints (19.6\%
vs.\
25.8\% success rate) 
The value of $p_\mathrm{KS}=0.081$, above the customary $\alpha=0.05$ significance threshold, would however tentatively support our use of
entropy-selected gaps in all situations where MT was used as a hint.

\paragraph{Comparing MT systems:}
Taking all MT systems
together, one can see that the success rate (58\%) is, as expected,
3~times 
larger than that
obtained without MT 
using the entropy-driven gap placing strategy (19\%)
and this difference is statistically significant. 
The homebrew Moses system is the least helpful (55.9\%), and Bing the
most helpful (62.6\%), but the only statistically significant
difference is between these two ($p_\mathrm{KS}=0.005$) and between
Bing and Systran ($p_\mathrm{KS}=0.044$). Even with 432 problems
solved for each system, MT systems were hard to distinguish by success
rate (Jordan et al.\ \shortcite{jordan17p} report clearer differences
between systems, but the paper does not clarify whether they are
running the same problems through all MT systems to ensure the independence
of their comparisons).

\begin{figure*}[ht]
  \centering
  \includegraphics[scale=0.45]{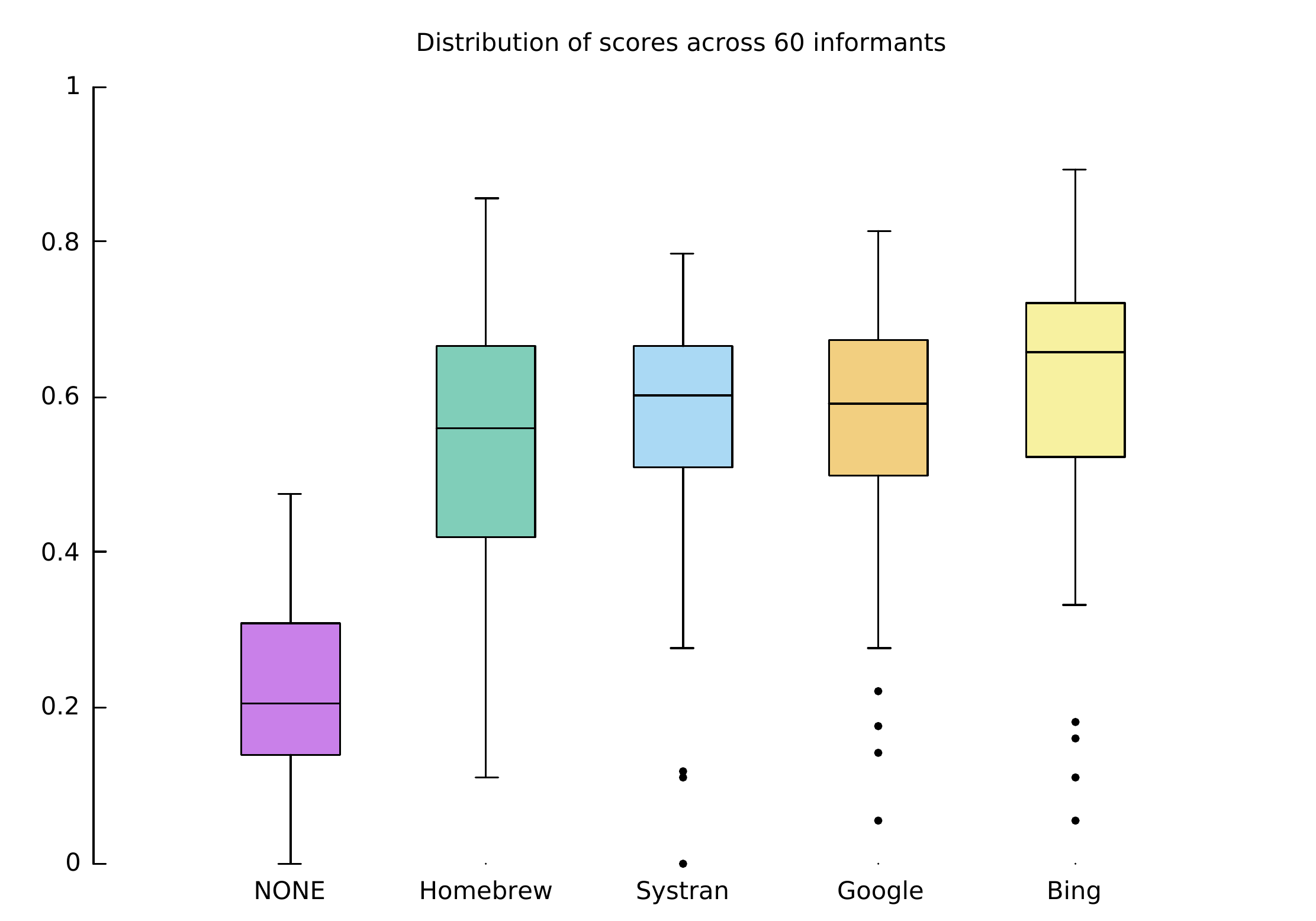}
  \caption{Box-and-whisker plots of the distribution of informant performance for each MT system.}
  \label{fig:bawp}
\end{figure*}

Figure~\ref{fig:bawp} shows box-and-whisker plots of the distribution of performance across all 60 informants for each MT system. The large overlap observed among the four MT systems illustrates how hard it is to simply average gap-filling scores to evaluate them.

Even if annotators are quite different, each one of them may still be consistent in the relative scores they give to different MT systems. Plotting the average score each informant gives to each MT system against their average score for all systems after removing four clearly outlying informants, Pearson correlations are only moderate (ranging between 0.47 and 0.73), and the slopes \(a_\mathrm{system}\) of line fits of the form \(\mathrm{score}(\mathrm{system})=a_\mathrm{system} \mathrm{score}(\mathrm{all})\) show the same ranking as average scores: 
\(a_\mathrm{homebrew}=0.95\), 
\(a_\mathrm{Systran}=0.97\), 
\(a_\mathrm{Google}=1.00\), 
\(a_\mathrm{Bing}=1.06\), but are very close to each other and their confidence intervals overlap substantially.

\paragraph{Effect of context:}
In half of the configurations with MT hints, a single machine-translated sentence
was shown; in the other half,
the whole machine-translated document was shown as a hint. The results 
indicate that extended context, instead of helping, seems to make the
task slightly more difficult (58.3\%
vs.\ 
59.5\% success rate), but 
differences are not statistically significant; therefore, GF scores
in Table~\ref{tb:quickcomp} are average scores obtained with and without
context. This supports evaluation through simpler GF tasks based on
single-sentence hints.

\paragraph{Effect of gap density:}
Gaps were punched with two different densities, 10\% and 20\%, to check
if a higher gap density would make the problem
harder. 
Contrary to intuition, the task becomes
easier when gap density is higher, and the result is
statistically significant
($p_\mathrm{KS}<0.001$).
This unexpected result is however easily explained as follows: problems
with 20\% gap density contain all of the high-entropy gaps present in
10\% problems, plus additional lower-entropy gaps, which are easier to
fill successfully, and therefore, the average success rate rises.  In
the no-hint situation, however, as shown in Table~\ref{tb:quickcomp},
higher densities would seem to make the problem harder, perhaps
because the only information available to fill the gaps comes from the
problem sentence itself, and 
higher gap densities substantially reduce the
%
number of available content words in the sentence.
However, the differences are not statistically significant. 

\paragraph{Gap density and MT evaluation:}
When comparing MT systems using only the 10\%
gap density problems, no differences are found to be statistically
significant. This means that for very hard gaps, systems would appear
to behave similarly. When selecting a value of 20\% for the gap
density (some easier gaps are included), Bing and Google do appear to
be significantly better than the homebrew Moses.

\paragraph{Inter-annotator agreement:} As 3 different informants filled
the gaps for exactly the same set of problems and configurations, with
20 such sets available, we studied the pairwise Pearson correlation
$r$ of their GF success in each of the 36 problems.\footnote{The usual
  Fleiss' kappa statistic cannot be applied here because the labels are not nominal or taken from a discrete set, but rather numerical success rates.} All values of $r$ were found to be
  positive, averaging around 0.58, a sign of rather good
  inter-annotator agreement.   After removing two outlying informants
  ($r<0.1$), results did not appreciably
  change. 


\paragraph{Allowing for synonyms:} The GF success scores reported thus far have been
computed by giving credit only to \emph{exact} matches. We have studied giving credit to \emph{synonyms} observed in informant work, namely to those appearing at least twice (in the work of all informants) that, according to one of the authors, preserved the meaning of the problem sentence, or were trivial spelling or case variations. A total of 124 frequent valid substitutions were considered.  As expected, GF
success rates (see table~\ref{tb:synonyms}) increase considerably%
, for example, from 22.7\% to 32.2\% for no hint, or from 58.9\% to 75.5\% for all systems averaged.
The relative ranking of MT systems is maintained;
the statistical significance of the homebrew Moses results versus Bing results
is maintained, and two additional statistically significant
differences appear: Google vs.\ homebrew Moses and Systran vs.\ homebrew Moses. The statistical significance of the effect of gap
density disappears when allowing for synonyms.
This indicates that it would be beneficial 
to assign credit to \emph{synonyms} if the necessary language resources are available or if further analysis of actual GF results is feasible.

\begin{table*}
  
    \begin{center}
    \begin{tabular}{c | c | c | c | c | c | c }

       & \multicolumn{3}{c|}{GF scores with synonyms} &  \multicolumn{3}{c}{GF scores without synonyms}\\ \hline
        System & Overall & 10\% & 20\% & Overall & 10\% & 20\% \\
      \hline
 	 Google & $0.757$ & $0.711$ & $0.776$ & $0.592$ & $0.565$ & $0.619$ \\     
      Bing &  $\mathbf{0.795}$ & $\mathbf{0.785}$ & $\mathbf{0.804}$ & $\mathbf{0.618}$ & $\mathbf{0.595}$& $\mathbf{0.640}$ \\ 
      Homebrew & $0.704$ & $0.711$ & $0.697$ & $0.550$ & $0.547$ & $0.553$ \\
      Systran & $0.765$ & $0.750$ & $0.781$  & $0.569$ & $0.544$ & $0.595$ \\\hline
      MT Average & $0.755$& $0.746$ & $0.765$ & $0.582$ & $0.563$ & $0.602$ \\\hline	
      No hint  (random) & $0.339$ & $0.379$ & $0.299$ & $0.258$   & $0.302$ & $0.213$ \\
      No hint  (entropy) & $0.306$ & $0.322$ & $0.290$ & $0.193$   & $0.191$ & $0.195$ \\
      No hint (average)  & $0.322$ & $0.350$ & $0.294$ & $0.225$   & $0.247$ & $0.204$ \\\hline

    \end{tabular}
    \end{center}
  
    \caption{Effect in success rates of allowing for synonyms in GF}
    \label{tb:synonyms}
 \end{table*}

\subsection{Correlation between GF and RCQ}

One of our main goals was to explore whether GF
would be able to reproduce the results of the
established method in the field, RCQ.
%
Table~\ref{tb:quickcomp} shows reasonable agreement between RCQ and GF scores: both
give the homebrew Moses system the worst 
score,
and 
commercial statistical systems (Bing and Google) 
get the best scores.
%
Also, 
as commonly 
found 
for subjective \emph{judgements} (for example, Callison-Burch et al.\
\shortcite{ccb06}), BLEU and NIST penalize the rule-based Systran system with
respect to the statistical homebrew system, while \emph{measurements}
of human performance do not, but the differences observed are however not
statistically significant.

 On the other hand, GF and RCQ scores
 assigned to specific (document, MT system) pairs show low
 correlation. This may be due to the scarcity of RCQ data (only one
 data point per document--MT system pair, as compared to of 12 data
 points for GF), or to the fact that, while RCQ takes the
 whole document into account, GF only looks at a specific sentence. 
 In addition, the RCQ tests and the sentence selected for GF for a given document may not directly correspond, i.e. the information required from the document to answer the RCQ tests may differ from the information required to fill the gaps in a given sentence. This happens because the comprehension questions may target different parts of the text and do not require the sentence selected by our GF approach. A natural follow up of this work is to use sentences for GF directly related to the RCQ tests.

\section{Concluding remarks}
\label{se:concluding}


We have compared two methods for the evaluation of MT in gisting
applications: the well-established method using reading comprehension questionnaires and %
an alternative method: gap filling.
While RCQ require the manual 
preparation of questionnaires for each document, 
and grading of answers to open questions, GF is cheaper, 
as it only needs reference translations for one or a few
sentences in each document and both questions and scores can be 
obtained automatically. GF is fast and easily crowdsourceable.

In GF, without a hint, we found that
entropy-selected gaps appear to be harder than random gaps. We therefore recommend using entropy-selected gaps to discourage guesswork and incentivize annotators to rely on the MT hints.
Providing the whole machine-translated document as a hint does not seem to help as
compared with providing only the machine-translated version of the problem
sentence. This would suggest the possibility of framing GF
evaluation around single sentences.

RCQ scores obtained using a machine-translated text range between
70\% and 95\% of the scores obtained using a professionally-translated
text. In GF, the presence of a machine-translated text clearly improves
performance (by about 
3~times).
Both results are a clear indication of
the usefulness of raw MT in gisting applications.

Both RCQ and GF 
rank a low-quality homebrew Moses system worst, but differ as
regards the best MT system, although differences are not always statistically significant. It would seem as if informants \emph{make do} with any MT system regardless of small differences in quality.
The discriminative power of
RCQ and GF evaluations 
is, however, quite low; this  
may be due to the scarcity
of data; if one expects that the collection of larger amounts of
human evaluation data (like the crowdsourced direct assessment (judgement) results described by Bojar et al.~\shortcite{bojar2016findings}) 
would
increase the discriminative power of the evaluation method, 
this would be much more feasible using GF, than the more costly RCQ.

 \paragraph{Acknowledgements:}
 Work supported by the Spanish government
 through project EFFORTUNE (TIN2015-69632-R) and through grant
 PRX16/00043 for MLF, and by the European Commission through project Health in my Language (H2020-ICT-2014-1, 644402). 
 CS is supported  by  the  EC  project  SIMPATICO  (H2020-EURO-6-2015, grant number 692819).
 We would like to thank Dr Ramon Ziai (University of T\"ubingen) for making the CREG corpus available and answering our questions.  

\bibliography{gapfilling.bib}

\begin{thebibliography}{36}
\expandafter\ifx\csname natexlab\endcsname\relax\def\natexlab#1{#1}\fi

\bibitem[{Ageeva et~al.(2015)Ageeva, Tyers, Forcada, and
  P{\'e}rez-Ortiz}]{ageeva2015evaluating}
Ekaterina Ageeva, Francis~M Tyers, Mikel~L Forcada, and Juan~Antonio
  P{\'e}rez-Ortiz. 2015.
\newblock Evaluating machine translation for assimilation via a gap-filling
  task.
\newblock In \emph{Proceedings of EAMT}, pages 137--144.

\bibitem[{Berka et~al.(2011)Berka, {\v{C}}ern{\'y}, and Bojar}]{berka2011quiz}
Jan Berka, Martin {\v{C}}ern{\'y}, and Ond{\v{r}}ej Bojar. 2011.
\newblock Quiz-based evaluation of machine translation.
\newblock \emph{The Prague Bulletin of Mathematical Linguistics}, 95:77--86.

\bibitem[{Bojar et~al.(2016)Bojar, Chatterjee, Federmann, Graham, Haddow, Huck,
  Yepes, Koehn, Logacheva, Monz et~al.}]{bojar2016findings}
Ond{\v{r}}ej Bojar, Rajen Chatterjee, Christian Federmann, Yvette Graham, Barry
  Haddow, Matthias Huck, Antonio~Jimeno Yepes, Philipp Koehn, Varvara
  Logacheva, Christof Monz, et~al. 2016.
\newblock Findings of the 2016 conference on machine translation.
\newblock In \emph{Proceedings of the First Conference on Machine Translation:
  Volume 2, Shared Task Papers}, volume~2, pages 131--198, Berlin, Germany.

\bibitem[{Bojar et~al.(2017)Bojar, Chatterjee, Federmann, Graham, Haddow,
  Huang, Huck, Koehn, Liu, Logacheva, Monz, Negri, Post, Rubino, Specia, and
  Turchi}]{bojar2017findings}
Ond\v{r}ej Bojar, Rajen Chatterjee, Christian Federmann, Yvette Graham, Barry
  Haddow, Shujian Huang, Matthias Huck, Philipp Koehn, Qun Liu, Varvara
  Logacheva, Christof Monz, Matteo Negri, Matt Post, Raphael Rubino, Lucia
  Specia, and Marco Turchi. 2017.
\newblock Findings of the 2017 conference on machine translation.
\newblock In \emph{Proceedings of the Second Conference on Machine Translation:
  Volume 2, Shared Task Papers}, volume~2, pages 169--214, Copenhagen, Denmark.

\bibitem[{Bojar et~al.(2015)Bojar, Chatterjee, Federmann, Haddow, Huck, Hokamp,
  Koehn, Logacheva, Monz, Negri, Post, Scarton, Specia, and
  Turchi}]{bojar2015findings}
Ond\v{r}ej Bojar, Rajen Chatterjee, Christian Federmann, Barry Haddow, Matthias
  Huck, Chris Hokamp, Philipp Koehn, Varvara Logacheva, Christof Monz, Matteo
  Negri, Matt Post, Carolina Scarton, Lucia Specia, and Marco Turchi. 2015.
\newblock Findings of the 2015 workshop on statistical machine translation.
\newblock In \emph{Proceedings of the Tenth Workshop on Statistical Machine
  Translation}, pages 1--46, Lisbon, Portugal. Association for Computational
  Linguistics.

\bibitem[{Callison-Burch et~al.(2006)Callison-Burch, Osborne, and
  Koehn}]{ccb06}
Chris Callison-Burch, Miles Osborne, and Philipp Koehn. 2006.
\newblock Re-evaluating the role of {BLEU} in machine translation research.
\newblock In \emph{EACL}, volume~6, pages 249--256.

\bibitem[{Castilho and O'Brien(2016)}]{Castilho2016}
Sheila Castilho and Sharon O'Brien. 2016.
\newblock {Evaluating the Impact of Light Post-Editing on Usability}.
\newblock In \emph{The Tenth International Conference on Language Resources and
  Evaluation}, pages 310--316, Portoro\v{z}, Slovenia.

\bibitem[{Castilho et~al.(2014)Castilho, O'Brien, Alves, and
  O'Brien}]{castilho2014does}
Sheila Castilho, Sharon O'Brien, Fabio Alves, and Morgan O'Brien. 2014.
\newblock Does post-editing increase usability? a study with {B}razilian
  {P}ortuguese as target language.
\newblock In \emph{Proceedings of the 17th Annual conference of the European
  Association for Machine translation, EAMT 2014}, pages 183--190. European
  Association for Machine Translation.

\bibitem[{Crook and Bishop(1965)}]{crook65tr}
M~Crook and H~Bishop. 1965.
\newblock Evaluation of machine translation, final report.
\newblock Technical report, Institute for Psychological Research, Tufts
  University, Medford, MA.

\bibitem[{Doherty and O'Brien(2009)}]{Doherty2009}
Stephen Doherty and Sharon O'Brien. 2009.
\newblock Can {MT} output be evaluated through eye tracking?
\newblock In \emph{The 12th Machine Translation Summit}, pages 214--221,
  Ottawa, Canada.

\bibitem[{Doherty and O'Brien(2014)}]{Doherty2014}
Stephen Doherty and Sharon O'Brien. 2014.
\newblock {Assessing the Usability of Raw Machine Translated Output: A
  User-Centred Study using Eye Tracking}.
\newblock \emph{International Journal of Human-Computer Interaction},
  30(1):40--51.

\bibitem[{Doherty et~al.(2010)Doherty, O'Brien, and Carl}]{Doherty2010}
Stephen Doherty, Sharon O'Brien, and Michael Carl. 2010.
\newblock Eye tracking as an automatic {MT} evaluation technique.
\newblock \emph{Machine Translation}, 24:1--13.

\bibitem[{Federmann(2012)}]{mtm12_appraise}
Christian Federmann. 2012.
\newblock Appraise: An open-source toolkit for manual evaluation of machine
  translation output.
\newblock \emph{The Prague Bulletin of Mathematical Linguistics}, 98:25--35.

\bibitem[{Fuji et~al.(2001)Fuji, Hatanaka, Ito, Kamei, Kumai, Sukehiro,
  Yoshimi, and Isahara}]{Fuji2001}
M.~Fuji, N.~Hatanaka, E.~Ito, S.~Kamei, H.~Kumai, T.~Sukehiro, T.~Yoshimi, and
  H.~Isahara. 2001.
\newblock {Evaluation Method for Determining Groups of Users Who Find {MT}
  ``Useful"}.
\newblock In \emph{The Eightth Machine Translation Summit}, pages 103--108,
  Santiago de Compostela, Spain.

\bibitem[{Fuji(1999)}]{Fuji1999}
Masaru Fuji. 1999.
\newblock Evaluation experiment for reading comprehension of machine
  translation outputs.
\newblock In \emph{The Seventh Machine Translation Summit}, pages 285--289,
  Singapore, Singapore.

\bibitem[{Heafield(2011)}]{heafield2011kenlm}
Kenneth Heafield. 2011.
\newblock {KenLM}: Faster and smaller language model queries.
\newblock In \emph{Proceedings of the Sixth Workshop on Statistical Machine
  Translation}, pages 187--197. Association for Computational Linguistics.

\bibitem[{Jones et~al.(2005{\natexlab{a}})Jones, Gibson, Shen, Granoien,
  Herzog, Reynolds, and Weinstein}]{jones2005measuring}
Douglas Jones, Edward Gibson, Wade Shen, Neil Granoien, Martha Herzog, Douglas
  Reynolds, and Clifford Weinstein. 2005{\natexlab{a}}.
\newblock Measuring human readability of machine generated text: three case
  studies in speech recognition and machine translation.
\newblock In \emph{Proceedings.(ICASSP'05). IEEE International Conference on
  Acoustics, Speech, and Signal Processing, 2005.}, volume~5, pages
  v:1009--v:1012. IEEE.

\bibitem[{Jones et~al.(2007)Jones, Herzog, Ibrahim, Jairam, Shen, Gibson, and
  Emonts}]{jones2007ilr}
Douglas Jones, Martha Herzog, Hussny Ibrahim, Arvind Jairam, Wade Shen, Edward
  Gibson, and Michael Emonts. 2007.
\newblock {ILR}-based {MT} comprehension test with multi-level questions.
\newblock In \emph{Human Language Technologies 2007: The Conference of the
  North American Chapter of the Association for Computational Linguistics;
  Companion Volume, Short Papers}, pages 77--80. Association for Computational
  Linguistics.

\bibitem[{Jones et~al.(2009)Jones, Shen, and Herzog}]{jones2009machine}
Douglas Jones, Wade Shen, and Martha Herzog. 2009.
\newblock Machine translation for government applications.
\newblock \emph{Lincoln Laboratory Journal}, 18(1).

\bibitem[{Jones et~al.(2005{\natexlab{b}})Jones, Gibson, Shen, Granoien,
  Herzog, Reynolds, and Weinstein}]{Jones2005a}
Douglas~A. Jones, Edward Gibson, Wade Shen, Neil Granoien, Martha Herzog,
  Douglas Reynolds, and Clifford Weinstein. 2005{\natexlab{b}}.
\newblock {{Measuring Translation Quality by Testing English Speakers with a
  New Defense Language Proficiency Test for Arabic}}.
\newblock In \emph{The International Conference on Intelligence Analysis},
  McLean, VA.

\bibitem[{Jordan-Núñez et~al.(2017)Jordan-Núñez, Forcada, and
  Clua}]{jordan17p}
Kenneth Jordan-Núñez, Mikel~L. Forcada, and Esteve Clua. 2017.
\newblock Usefulness of {MT} output for comprehension --- an analysis from the
  point of view of linguistic intercomprehension.
\newblock In \emph{Proceedings of MT Summit XVI}, volume 1. Research Track,
  pages 241--253.

\bibitem[{Klerke et~al.(2015)Klerke, Castilho, Barrett, and
  S\o{gaard}}]{Klerke2015}
Sigrid Klerke, Sheila Castilho, Maria Barrett, and Anders S\o{gaard}. 2015.
\newblock {Reading metrics for estimating task efficiency with {MT} output}.
\newblock In \emph{The Sixth Workshop on Cognitive Aspects of Computational
  Language Learning}, pages 6--13, Lisbon, Portugal.

\bibitem[{Koehn et~al.(2007)Koehn, Hoang, Birch, Callison-Burch, Federico,
  Bertoldi, Cowan, Shen, Moran, Zens, Dyer, Bojar, Constantin, and
  Herbst}]{Kohen2007}
Philipp Koehn, Hieu Hoang, Alexandra Birch, Chris Callison-Burch, Marcello
  Federico, Nicola Bertoldi, Brooke Cowan, Wade Shen, Christine Moran, Richard
  Zens, Chris Dyer, Ond\v{r}ej Bojar, Alexandra Constantin, and Evan Herbst.
  2007.
\newblock Moses: Open source toolkit for statistical machine translation.
\newblock In \emph{The Annual Meeting of the Association for Computational
  Linguistics, demonstration session}, Prague, Czech Republic.

\bibitem[{Meurers et~al.(2011)Meurers, Ott, and Kopp}]{Meurers2011}
Ramon~Ziai Meurers, Niels Ott, and Janina Kopp. 2011.
\newblock {{Evaluating Answers to Reading Comprehension Questions in Context:
  Results for German and the Role of Information Structure}}.
\newblock In \emph{TextInfer 2011 Workshop on Textual Entailment}, pages 1--9,
  Edinburgh, UK.

\bibitem[{O'Regan and Forcada(2013)}]{oregan2013p}
Jim O'Regan and Mikel~L. Forcada. 2013.
\newblock Peeking through the language barrier: the development of a
  free/open-source gisting system for {B}asque to {E}nglish based on
  apertium.org.
\newblock \emph{Procesamiento del Lenguaje Natural}, pages 15--22.

\bibitem[{Ott et~al.(2012)Ott, Ziai, and Meurers}]{Ott2012}
Niels Ott, Ramon Ziai, and Detmar Meurers. 2012.
\newblock Creation and analysis of a reading comprehension exercise corpus:
  Towards evaluating meaning in context.
\newblock In T.~Schmidt and K.~Worner, editors, \emph{Multilingual Corpora and
  Multilingual Corpus Analysis}, Hamburg Studies on Multilingualism (Book 14),
  pages 47--69. John Benjamins Publishing Company, Amsterdam, The Netherlands.

\bibitem[{Sager(1993)}]{sager93b}
Juan~C. Sager. 1993.
\newblock \emph{Language engineering and translation: consequences of
  automation}.
\newblock Benjamins, Amsterdam.

\bibitem[{Sajjad et~al.(2016)Sajjad, Guzman, Durrani, Bouamor, Abdelali,
  Teminkova, and Vogel}]{Sajjad2016}
Hassan Sajjad, Francisco Guzman, Nadir Durrani, Houda Bouamor, Ahmed Abdelali,
  Irina Teminkova, and Stephan Vogel. 2016.
\newblock {Eyes Don't Lie: Predicting Machine Translation Quality Using Eye
  Movement}.
\newblock In \emph{The 2016 Conference of the North American Chapter of the
  Association for Computational Linguistics: Human Language Technologies},
  pages 1082--1088, San Diego, CA.

\bibitem[{Scarton and Specia(2016)}]{scarton16p}
Carolina Scarton and Lucia Specia. 2016.
\newblock A reading comprehension corpus for machine translation evaluation.
\newblock In \emph{Proceedings of the Tenth International Conference on
  Language Resources and Evaluation (LREC 2016)}, Paris, France. European
  Language Resources Association (ELRA).

\bibitem[{Sinaiko and Klare(1972)}]{sinaiko72j}
H.~Wallace Sinaiko and George~R. Klare. 1972.
\newblock Further experiments in language translation.
\newblock \emph{International Journal of Applied Linguistics}, 15:1--29.

\bibitem[{Somers and Wild(2000)}]{somers00p}
Harold Somers and Elizabeth Wild. 2000.
\newblock Evaluating machine translation: the cloze procedure revisited.
\newblock In \emph{Translating and the Computer 22: Proceedings of the
  Twenty-second International Conference on Translating and the Computer}.

\bibitem[{Stymne et~al.(2012)Stymne, Danielsson, Bremin, Hu, Karlsson,
  Lillkull, and Wester}]{Stymne2012}
Sara Stymne, Henrik Danielsson, Sofia Bremin, Hongzhan Hu, Johanna Karlsson,
  Anna~Prytz Lillkull, and Martin Wester. 2012.
\newblock {Eye Tracking as a Tool for Machine Translation Error Analysis}.
\newblock In \emph{The 8th International Conference on Language Resources and
  Evaluation}, pages 1121--1126, Istanbul, Turkey.

\bibitem[{Taylor(1953)}]{taylor1953cloze}
Wilson~L Taylor. 1953.
\newblock ``{C}loze procedure'': a new tool for measuring readability.
\newblock \emph{Journalism Bulletin}, 30(4):415--433.

\bibitem[{Tomita et~al.(1993)Tomita, Masako, Tsutsumi, Matsumura, and
  Yoshikawa}]{Tomita1993}
Masaru Tomita, Shirai Masako, Junya Tsutsumi, Miki Matsumura, and Yuki
  Yoshikawa. 1993.
\newblock Evaluation of {MT} systems by {TOEFL}.
\newblock In \emph{The Fifth International Conference on Theoretical and
  Methodological Issues in Machine Translation}, pages 252--265, Kyoto, Japan.

\bibitem[{Trosterud and Unhammer(2012)}]{trosterud2012evaluating}
Trond Trosterud and Kevin~Brubeck Unhammer. 2012.
\newblock Evaluating {N}orth {S}{\'a}mi to {N}orwegian assimilation {RBMT}.
\newblock In \emph{Proceedings of the Third International Workshop on
  Free/Open-Source Rule-Based Machine Translation (FreeRBMT 2012)}.

\bibitem[{Weiss and Ahrenberg(2012)}]{weiss2012error}
Sandra Weiss and Lars Ahrenberg. 2012.
\newblock Error profiling for evaluation of machine-translated text: a
  {P}olish--{E}nglish case study.
\newblock In \emph{LREC}, pages 1764--1770.

\end{thebibliography}
\bibliographystyle{acl_natbib_nourl}
\appendix
\section{Supplemental material} 
\label{app:supplemental}
\paragraph{Raw gap-filling results} for 2159 problems,\footnote{Should have been $2160=36\times 60$, but data for one specific document, informant and configuration, was lost due to a bug in the Appraise system.} 60 informants,
36 documents, and 20 configurations, are available for download at the
following address: \url{http://www.dlsi.ua.es/~mlf/wmt2018/raw-gap-filling-results.csv}.

\paragraph{Raw reading comprehension test results} for 36 documents, four different MT systems (Google, Bing, Moses and Systran) and one human reference are available, totalling 180 documents. Each document was assessed by one test taker. The markings for questions available in each document and the final document scores used in this paper (namely simple, weighted or literal) are available for download at: 
\url{http://www.dlsi.ua.es/~mlf/wmt2018/raw-reading-comprehension-results.csv}.

\end{document}